\def\year{2022}\relax
\documentclass[letterpaper]{article} 
\usepackage{aaai22}  
\usepackage{times}  
\usepackage{helvet}  
\usepackage{courier}  
\usepackage[hyphens]{url}  
\usepackage{graphicx} 
\usepackage{natbib}  
\usepackage{caption} 
\DeclareCaptionStyle{ruled}{labelfont=normalfont,labelsep=colon,strut=off} 
\usepackage{subcaption}
\usepackage{amsmath}
\usepackage{amsfonts}
\usepackage{booktabs,siunitx}
\usepackage{algorithm}
\usepackage{algorithmic}

\usepackage{newfloat}
\usepackage{listings}
\title{Graph network for learning bi-directional physics}
\author{
    Sakthi Kumar Arul Prakash\textsuperscript{\rm 1}, Conrad Tucker\textsuperscript{{\rm 1},{\rm 2},{\rm 3},{\rm 4},{\rm 5}}
}
\affiliations{
    \textsuperscript{\rm 1}Department of Mechanical Engineering, Carnegie Mellon University \\
    \textsuperscript{\rm 2}Department of Machine Learning, Carnegie Mellon University \\
    \textsuperscript{\rm 3}The Robotics Institute, Carnegie Mellon University \\
    \textsuperscript{\rm 4}CyLab Security and Privacy Institute, Carnegie Mellon University \\
    \textsuperscript{\rm 5}Department of Biomedical Engineering, Carnegie Mellon University \\

    5000 Forbes Avenue, Pittsburgh, Pennsylvania 15213 \\
    sarulpra@andrew.cmu.edu, conradt@andrew.cmu.edu
}

\usepackage{bibentry}

\begin{document}

\maketitle

\begin{abstract}
In this work, we propose an end-to-end graph network that learns forward and inverse models of particle-based physics using interpretable inductive biases. Physics-informed neural networks are often engineered to solve specific problems through problem-specific regularization and loss functions. Such explicit learning biases the network to learn data specific patterns and may require a change in the loss function or neural network architecture hereby limiting their generalizabiliy. Our graph network is implicitly biased by learning to solve several tasks, thereby sharing representations between tasks in order to learn the forward dynamics as well as infer the probability distribution of unknown particle specific properties. We evaluate our approach on one-step next state prediction tasks across diverse datasets. Our comparison against related data-driven physics learning approaches reveals that our model is able to predict the forward dynamics with at least an order of magnitude higher accuracy. We also show that our approach is able to recover multi-modal probability distributions of unknown physical parameters.  
\end{abstract}

\section{Introduction}

Depending on the complexity of the problem, the availability of an accurate physics simulator and the amount of data, a model that solves either or both of the above problems can include domain knowledge in various forms. Physics-informed neural networks explicitly include domain knowledge in the form of problem specific loss functions, often requiring diverse data obtained by varying all the properties that govern the dynamics of the problems which may be difficult to obtain~\cite{Raissi2019,Obiols-Sales,Zeng2020}. Data-driven models do not use custom loss functions and may implicitly induce biases to discover the dynamics of the system~\cite{Battaglia2013,Battaglia2016, Sanchez-Gonzalez2018a}. However, such models require more data than their explicit learning counter-parts in order to achieve comparable accuracy. Finally, there are hybrid models that combine explicit and implicit biasing strategies such as fusing differentiable physics simulators with neural networks, or incorporating physics intuition as implicit bias.

This work introduces a hybrid graph network that jointly learns the forward dynamics and also infers the unknown physical properties. We decompose Newtonian physics models into deterministic and probabilistic sub-components. We represent each of the sub-components as learnable graph networks; a \textit{Forward model} using deterministic graph auto-encoders/decoders and an \textit{Inverse model} using a differentiable probabilistic graphical model. The sub-components are composed together to form a single graph network. Together, they learn a shared encoded representation of the particles and their interactions using a supervised learning objective. Apart from predicting the next state and particle specific unknown physical parameter, we explore other implicit biasing strategies in this paper such as particle classification and contact prediction. We hypothesize that composing inverse and forward models along with implicit learning tasks, induces a structural bias in the overall learning of problem dynamics, thereby avoiding the need for problem specific loss functions while using orders of magnitude fewer samples. We validate our model by evaluating on several particle systems that differ in the physics being learned.

\section{Approach}
We propose a model that improves the expressiveness of current Graph Network architectures as learnable simulators that predict the next state of a system of particles and also infers the probability distribution of unknown physical parameters. 
\subsection{Learned Simulator Overview}
Let ${\Omega}$ be a continuous real space of all possible states a system of particles~($\textbf{P}$) can exist in any given time $t$. Consequently, since $\textbf{P}^t \in \Omega$, the current state~($\textbf{P}_i^t \in \textbf{P}^t$) of each particle~$i$ describes a position vector~($\textbf{X}_{i}^{t} \in \mathbb{R}^N$) and a velocity vector~($\dot{\textbf{X}}_{i}^{t} \in \mathbb{R}^{N}$) such that $\textbf{P}_i^t = [\textbf{X}_{i}^{t}, \dot{\textbf{X}}_{i}^{t}, \dot{\textbf{X}}_{i}^{t-1} \dots ,\dot{\textbf{X}}_{i}^{t-k}]$, where $N$ represents the dimensionality of the problem and $k$ is a hyper-parameter that denotes the number of previous velocity vectors to consider in the state representation of each particle.
Since we adopt a particle-based representation of the physical system, we consider particles as nodes~($V$) in a graph~($\mathcal{G}$) such that every particle has a local first-order neighborhood as well as hops of higher-order neighborhoods. Such an explicit graph structure further imposes inductive bias that minimizes the dependence of distant particles from target particles. Hence, every system state~$\textbf{P}^t \in \Omega$ has a corresponding adjacency matrix~($\textbf{A}^t$) which is a matrix representation of $\mathcal{G}^t = (V,E^t)$. The number of particles~$|V|$ stays a constant throughout the simulation since we assume that there is no new addition or removal of particles. In contrast, $E^t$-- the edges of the graph, may vary as a function of time and hence the superscript~$t$.

\begin{figure*}[!ht]
    \centering
    \includegraphics[width=0.92\textwidth]{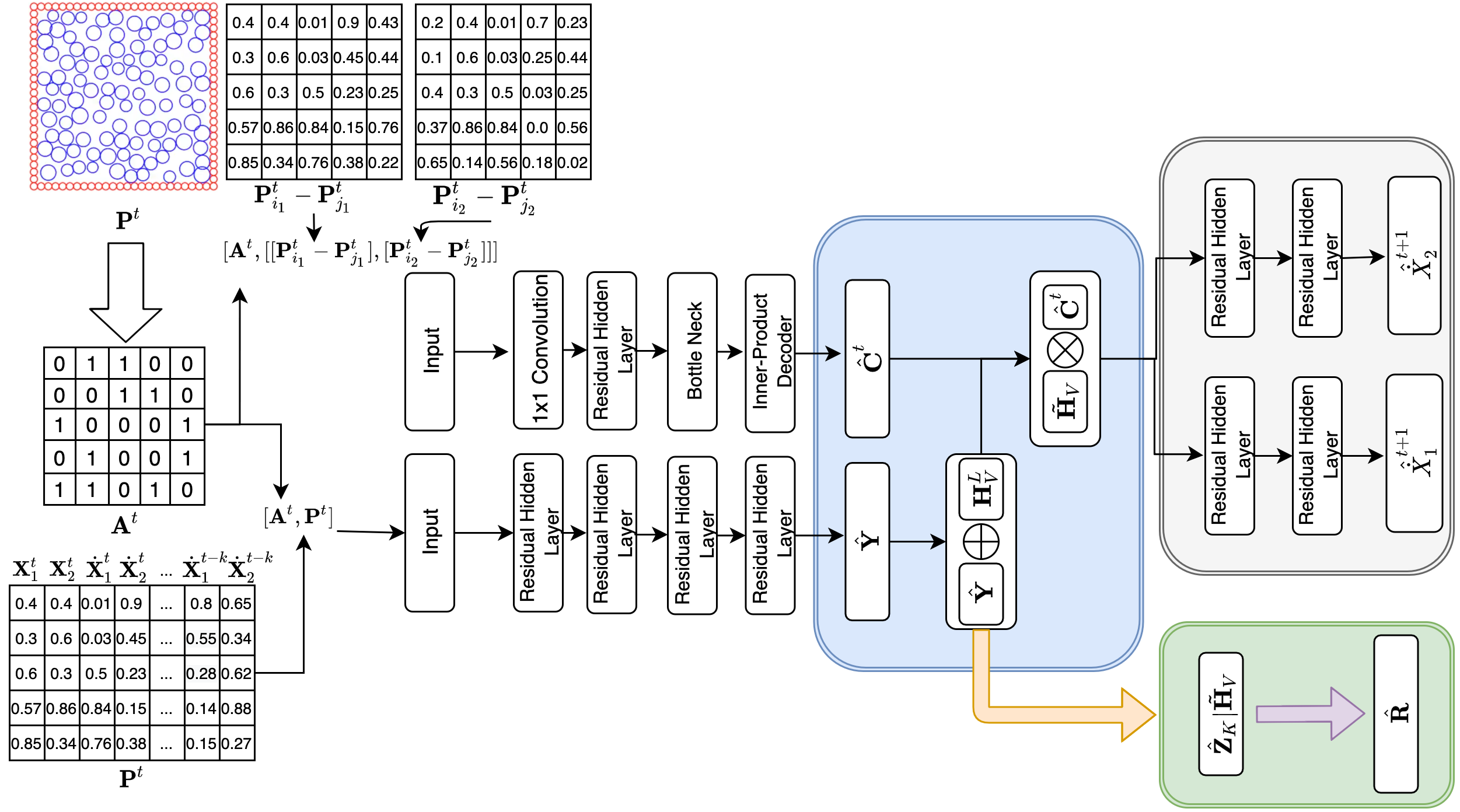}
    \caption{Proposed Model: Our approach solves a multi-task learning problem that seeks to identify the particle type~($Y_i$) and contact matrix~($\hat{\textbf{C}}$) to predict the next state's target velocity or acceleration in a 2D scenario while also approximating the unknown inverse parameter~(eg.,Mass). Blue shading indicates common processor for forward simulation decoder and inverse model, Grey shading depicts forward simulation decoder, while green shading depicts the inverse model decoder.}
    \label{fig:proposed_model}
\end{figure*}

A forward simulator $S:\textbf{P}^t\mapsto \textbf{P}^{t+1}$, the next state is predicted as, $\hat{\textbf{P}}^{t+1} = S(\textbf{P}^t)$. We are interested in learning the dynamics of the system using a parameterized function approximator~($S_{\theta}$) as a surrogate to an actual physics simulator~($S$) which can typically be complex. In addition to learning a forward simulator, we learn unknown physical properties of particles such as mass by using a parameterizable latent variable model~$f_{\phi}$ that approximates the probability density function of a random variable~($Z$) that represents the unknown property of interest pertaining to all particle types from the simulation distribution. In this work, we assume that each particle has only one unknown physical parameter that needs to be estimated. The learnable parameter~${\phi}$ is optimized using a variational inference training objective. Given multiple learning tasks that are related, we frame the problem to minimize multi-task learning objectives using the knowledge contained from all the tasks. $S_{\theta}$ corresponds to four steps-- Encoding, Processing, Inverse parameter estimation and Decoding. The architecture of the proposed learned simulator is shown in~Fig.~\ref{fig:proposed_model}.
\subsection{Encoder definition and design}
Following the established notations, we define a particle Encoder~$\mathcal{E}_V: [\textbf{A},\textbf{P}]^t \mapsto \textbf{H}_V$ where $\textbf{H}_V$ is an encoded particle representation. Next, we define a relation Encoder~$\mathcal{E}_E: [\textbf{A}, {\textbf{P}_i-\textbf{P}_j}]^{t} \mapsto \textbf{H}_E$ where $(\textbf{P}_i-\textbf{P}_j)^{t}$ represents the relative position and velocity information between particles, and $\textbf{H}_E$ is an encoded edge representation of the system of particles. The relative matrix~$({\textbf{P}_i - \textbf{P}_j})^t$ is a tensor of size $N\times(k+1)\times|V|\times|V|$. The computation graph of the absolute state encoder~($\mathcal{E}_V$) is made up of raw-residual graph convolution layers that have shown to prevent information smoothing~\cite{DBLP:journals/corr/abs-1909-05729} in deeper layers while capturing node specific graph structural features.

 While prior approaches have used particle type as input to the Graph Networks, we consider learning particle types as one of the tasks. By mapping the input state to particle type, the encoding layers learn discriminative particle specific representations which summarize a patch of the graph centered around a particle~$i$. Such discriminative representations have helped downstream tasks such as node classification~\cite{DBLP:conf/iclr/KipfW17,Prakash2021} which is also what $\mathcal{E}_V$ computes. A single layer of $\mathcal{E}_V$ can mathematically be expressed as follows,
\begin{equation}
\begin{split}
    \textbf{H}^{l+1}_{V} &= g(\textbf{H}^{l}_{V}) \\
    &= \sigma(\hat{\textbf{A}}^{t}\textbf{H}^{l}_{V}\textbf{W}^{l}_{V} + \hat{\textbf{A}}^{t}\textbf{H}^{l}_{V})
\end{split}
\end{equation}
where, $g(.)$ is a learnable raw-residual graph convolution layer, $\sigma(.)$ is a non-linear activation function $\textbf{H}_V^{l=0} = \textbf{P}^t$, $\textbf{H}_V^{l}$ represents $l$th hidden layer, such that $l = [1 \ldots L]$ and $\hat{\textbf{A}}^{t} = \textbf{A}^{t} + \textbf{I}$, where $\textbf{I}$ is an identity matrix.

Concurrently, while the computation graph of the relative state encoder~($\mathcal{E}_E$) is made up of similar hidden layers, it captures the edge specific graph structural features~\cite{Wang2019} that encode complex information such as contact, collision, friction, etc. We assume that complex phenomenon such as contact, collision or friction between particles can be summarized as a $|V|\times|V|$ matrix that denotes whether a direct contact has occurred using a binary matrix in the case of direct contact and a weighted matrix in case of indirect contact such as the influence of gravitational force. Hence, in order to create dependence between the other channels, we apply a $1\times1$~convolution layer on $(\textbf{P}_i - \textbf{P}_j)^t$ to transform it to a $|V|\times|V|$ matrix that is compatible with $g(.)$ in order to compute~$\textbf{H}^{1}_{E}$.
\subsection{Processor definition and design}
The processor is an integral component in the proposed learnable simulator that biases edge and particle features through several tasks such as particle classification, contact detection, composing a learned prior for the inverse model and communicating with the forward simulation decoder. While we do not provide a theoretical description on the kinds of tasks that can be performed, we consider some example tasks and discuss their contribution in implicit learning. We briefly describe several tasks the processor performs in this paper. It takes the output of the encoder~($\mathcal{E}_V$) that computes particle specific latent graph features~($\textbf{H}^{L}_{V}$) as input to predict the particle type~($\hat{\textbf{Y}}$) for all particles in the system as given by Eq.(2)
\begin{equation}
    \hat{\textbf{Y}} = \alpha(\textbf{H}^{L}_{V}\textbf{W}^{L})
\end{equation}
where, $\alpha$ can be a categorical softmax activation in case of several particle types or a sigmoid activation function in case of two particle types. The processor trains the classifier to minimize the following cross-entropy loss, $\mathcal{L}_{classification} = -\sum_{i \in \mathcal{V}}[\textbf{Y}_{i}\ln{\hat{\textbf{Y}}_{i}}]$ in a supervised setting. Once the particle type has been identified, the current latent particle representations~($\textbf{H}^{L}_{V}$) are updated as $\tilde{\textbf{H}}_{V} = \textbf{H}^{L}_{V}||\hat{\textbf{Y}}$, where $||$ indicates concatenation.  The updated representation~$\tilde{\textbf{H}}_{V}$ will also serve as input to the inverse model. 

As an example of another possible task, the processor takes the output of the encoder~($\mathcal{E}_E^{L}$) as input to a Decoder~$\mathcal{D}_{\theta_E}: \textbf{H}_E \mapsto \hat{\textbf{C}}$ where $\hat{\textbf{C}}$ is a contact matrix that estimates contact magnitude between particles at time~$t$. The contact matrix~$\hat{\textbf{C}}^t$ induces an inductive bias on the feature matrix~$\tilde{\textbf{H}}_{V}$ such that the updated representations capture absolute and relative information. $\mathcal{D}_{\theta_E}$ is a simple inner-product decoder that can be represented as follows,
\begin{equation}
    \hat{\textbf{C}}^t = \textbf{H}^{L}_{E}(\textbf{H}^{L}_{E})^T
\end{equation}
The contact detector module within the processor is trained to reconstruct a matrix by minimizing the following mean squared error,
\begin{equation}
    \mathcal{L}_{collision} = \frac{1}{|V|}\sum_{i=1}^{|V|}{||\textbf{C}_i^t-\hat{\textbf{C}}_i^t||}_2^2
\end{equation}
The purpose of~$\mathcal{E}_E$ and $\mathcal{D}_E$ is to emulate contact estimation that several Newtonian physics models possess~\cite{Cant2008,Glatt2019} such as collision detection in the case of direct contact models and contact magnitudes in the case of indirect models. Similarly, once particle contacts have been estimated, the processor concatenates $\tilde{\textbf{H}}_{V}$ and the contact mask~$\hat{\textbf{C}}^t$ as 
$\tilde{\textbf{H}}_{CV} = \hat{\textbf{C}}^t||{\textbf{H}}_{V}$. 
\subsection{Decoder definition and design}
There are $N$ decoders~$\mathcal{D}_{\theta}$ that each comprise of raw-residual linear layers that maps the processor's output~$\tilde{\textbf{H}}_{CV}$ to~$\hat{\dot{X}}_{i_{1}}^{t+1}$ and $\hat{\dot{X}}_{i_{2}}^{t+1}$ in the case when $N=2$ for a particle~$i$ as shown in the grey block in Fig.~\ref{fig:proposed_model}. This is expressed mathematically as follows,
\begin{equation}
    \hat{\dot{\textbf{X}}}_{(.)}^{t+1} = \mathcal{D}_{\theta}(\sigma(\tilde{\textbf{H}}_{CV}\textbf{W}_{{Dec}_{(.)}}^{l=0})\textbf{W}_{Dec_{(.)}}^{l=1})
\end{equation}
where, $(.)$ denotes $N^{th}$ dimensional index and $\textbf{W}_{{Dec}_{(.)1}}$ and $\textbf{W}_{{Dec}_{(.)2}}$ are learnable weight matrices of a two layer decoder. Since the output of~$\mathcal{D}_{\theta}$ is a regression output, we reconstruct the target velocities or acceleration~(depending on problem) by minimizing the following mean squared error loss function, $\mathcal{L}_{Dec}=\frac{1}{|V|}\sum_{i=1}^{|V|}||\dot{\textbf{X}}_{i(.)}^{t+1} - \hat{\dot{\textbf{X}}}_{i(.)}^{t+1}||_2^2$.
\subsection{Inverse model definition and design}
We now consider the scenario of a single unknown physical parameter pertaining to a system of particles that we denote using a continuous random variable~$M$, whose unknown probability density function~($p_M(m)$) we are interested to estimate, given~$\tilde{\textbf{H}}_{V}$. Further, we would like to train a parameterizable function~$f_\phi$ in a supervised setting where we assume that we have access to the absolute physical parameter values of some particles and also in an unsupervised setting where we assume no knowledge of the physical parameter values. Hence our objective can either be seen as minimizing~$D_{KL}(f_{\phi}||\tilde{p}_M)$ or $D_{KL}(\tilde{p}_M||f_{\phi})$, where the former and latter can be recognised as the reverse and forward KL divergences respectively. We know that while one learning objective penalises mode-missing~(reverse KL), the other penalises mode-covering~(forward KL). However, from an inverse modelling perspective, mode dropping does not provide a complete picture of the distribution of a physical parameter of interest. Additionally, reverse KL objective requires a tractable likelihood function as a workaround to prevent mode missing, but such a function is usually unavailable. On the other hand, while the forward KL objective may present an over-smoothed picture of the distribution, a potential work-around is to use an expressive model to approximate the distribution using the parameter values of known particle samples.

To that end, we frame the density estimation of the posterior predictive or the unknown physical parameter as a conditional density estimation problem using a parameterizable flow model~$f_\phi(.)$ that comprises of ~$K$ sequential flow layers. The normalizing flows approach~\cite{DBLP:conf/icml/RezendeM15,DBLP:conf/nips/KingmaSJCCSW16} provides a tractable way of computing the posterior log-likelihood of~$\tilde{p}_M$ as given by~Eq.~(7) 
\begin{equation}
    \log {p}(z_K) = \log {p}(z_0) - \sum_{k=1}^{K}\log\biggm\lvert\det\biggm(\frac{\partial f_{\phi}(z_{k-1})}{z_{k-1}}\biggm)\biggm\lvert
\end{equation}
where $z_0\sim \mathcal{N}(0,1)$, $z_K\sim f_\phi(z_{K-1})$, $\log {p}(z_K)$ is the log-likelihood approximation of $M$ and $\log {p}(z_0)$ is the log-likelihood approximation of a prior standard Gaussian distribution. However, instead of reversing the direction of the flows to estimate the density~\cite{Trippe2018}, we preserve the conventional direction of the flows in order to quickly sample the unknown physical parameter from the posterior. To accomplish this feat we use flow layers proposed by~\cite{Huang2018} that have shown to be expressive in recovering all modes of an unknown density function. While the flow direction implicitly fits a reverse KL divergence, the learning objective as given by Eq.~(8) minimizes the forward KL divergence objective as follows:
\begin{equation}
\begin{split}
    D_{KL}(\tilde{p}_M||p_{z_{K}}) &= \mathbb{E}_{m\sim\tilde{p}_M}[\log\tilde{p}_M] - \mathbb{E}_{m\sim\tilde{p}_M}[\log p(z_K)] \\
    &= \mathcal{H}_M + \mathbb{E}_{m\sim\tilde{p}_M}[-\log p(z_K)] \\
    &\approx \frac{1}{|V|}\sum_{i=1}^{|V|}-\log p(z_K)
\end{split}
\end{equation}
We sample~$|V|$ particle specific physical parameter values~($Z_{K}$) from the posterior and perform a bi-linear transformation with a learnable weight matrix~$\textbf{W}$ that captures pairwise interactions between the absolute parameter values of the samples using Eq.~(9), 
 \begin{equation}
     \textbf{H}_{Inv-Dec} = Z_{K}^T\textbf{W}_{Inv-Dec_{1}}Z_{K}
 \end{equation}
and are then further passed through several linear layers to predict the relative masses of all particles. Thus the final output of the decoder of the inverse model~$\mathcal{D}_{\theta_{inv}}:Z_{K} \mapsto \hat{\textbf{R}}$ where $\hat{\textbf{R}}$ is a reconstruction of the relative physical parameter matrix~$\textbf{R}$ that can be expressed using the following expression,
\begin{equation}
    \hat{\textbf{R}} = \mathcal{D}_{\theta_{Inv}}((\sigma((Z_{K}^T\textbf{W}_{Inv_{1}}Z_{K}))\textbf{W}_{{Inv-Dec}_{2}})\textbf{W}_{Inv-Dec_{3}})
\end{equation}
We minimize the reconstruction loss~$\mathcal{L}_{Inv-Dec}=\frac{1}{|V|}\sum_{i=1}^{|V|}||\textbf{R}_{i} - \hat{\textbf{R}}_{i}||_2^2$ along with the inverse model encoder loss~$D_{KL}(\tilde{p}_M||p_{z_K})$ such that maximizes the following log-likelihood,
\begin{equation}
    \log p_\theta(\textbf{R}) = - D_{KL}(\tilde{p}_M||p_{z_K}) + \mathbb{E}_{\sim\textbf{R}}[\log p(\textbf{R}_i|z_K;\theta)]
\end{equation}
Finally, we put together the combined loss function from all the steps as follows:
\begin{equation}
    \mathcal{L}^{Total} = \mathcal{L}_{Inverse} + \mathcal{L}_{Dec} + \mathcal{L}_{collision} + \mathcal{L}_{classification}
\end{equation}
where, $\mathcal{L}_{Inverse} = -\log p_\theta(\textbf{R})$.
\section{Discussion}
In this section, we describe the datasets that were used to test our approach and discussion on key findings from a diverse set of experiments. In order to evaluate the rigorousness of the proposed approach, we keep a number of hyper-parameter choices such as the number of hidden layers, hidden neurons, etc. constant~(please refer to appendix for more details) and further compare with related work~\cite{Sanchez-Gonzalez2020,DBLP:conf/nips/CranmerSBXCSH20}. 
\subsection{Dataset}
Our datasets are simulations of common Newtonian dynamics that describes the dynamics of particles according to Newton’s law of motion. The motion of particles are predicted using their complex interactions with neighboring particles that cause a change in their position, velocity and accelerations. We evaluate our approach on simulations that involve direct contact namely, elastic rigid body collisions, and indirect contacts wherein particles are subject to spring and gravitational forces. The elastic rigid-body simulation was written using numpy and scipy libraries while the other simulations that deal with spring and gravitational particle interactions~\cite{DBLP:conf/nips/CranmerSBXCSH20} were written using the JAX library. The variable parameters of these analytic simulation models include the number of simulations, number of particles, time-steps and frame-rate~(step-size). Each simulation per analytic model is integrated over 1000 time-steps which is typically the simulation length used by related work such as~\cite{DBLP:conf/nips/CranmerSBXCSH20, Sanchez-Gonzalez2020}, and our dataset comprises of 10 simulations per analytic model. We set the number of particles = 200 in the case of 2D elastic rigid-body collisions and a standard size of 4 in the case of 2D spring and gravitational force simulations.
\subsection{Experiments and Discussion}
Simulators that employ explicit and semi-implicit methods of integration for updating particle states are error bounded based on step-size or frame rate. We evaluate the forward and inverse estimation performance of our proposed approach across varying frame rates. Frame-rate or also referred to as step-size is an important parameter in physics simulations since they directly influence the stability of the simulator as well as the accuracy of the simulator's estimations~\cite{Arnold2001, Lei2012}. We evaluate Cranmer et al.~\cite{DBLP:conf/nips/CranmerSBXCSH20} and Sanchez-Gaonzalez et al. models on  simulator specific hand-tuned frame rates. Further, we train and test their approaches on $10^6$ samples as reported by the authors in order to present a  fair comparison. We also investigate how our approach is able to estimate an unknown physical parameter's probability distribution- the \textit{mass} distribution, when the frame rate and the complexity of the distribution varies. We consider 5 frame rates ranging from 10-200 as a factor and compare our model performance on all the datasets and against related work.   
\subsubsection{Performance of forward model}
\begin{table*}[!h]
\centering
\caption{Mean one-step target velocity/acceleration prediction error}
\label{table:Forward Simulation Performance}
\scalebox{0.97}{
\begin{tabular}{|l|l|l|l|l|l|l|}
\hline
    \multicolumn{1}{|c|}{Datasets} &
    \multicolumn{1}{c|}{10 fps} &
    \multicolumn{1}{c|}{30 fps} &
    \multicolumn{1}{c|}{50 fps} &
    \multicolumn{1}{c|}{100 fps} &
    \multicolumn{1}{c|}{\cite{DBLP:conf/nips/CranmerSBXCSH20}} &
    \multicolumn{1}{c|}{\cite{Sanchez-Gonzalez2020}} \\
    \hline
    2D Elastic Collisions & \textbf{0.003} & \textbf{0.003} & \textbf{0.003} & 0.006  & 0.119 & 0.053  \\
    \hline
    2D Spring & \textbf{0.0007} & 0.093 & 0.0009 & 0.08 & 0.047 & 0.734  \\
    \hline
    2D Gravity~($\frac{1}{r^2}$) & 1.023 & 1.363 & 1.306 & \textbf{0.804} & 1.634 & 1.804  \\
    \hline
    2D Gravity~($\frac{1}{r}$) & 0.129 & 0.561 & 0.588 & 0.836 & \textbf{0.077} & 1.340 \\
    \hline
\end{tabular}}
\end{table*}
The graph network architectures proposed by Cranmer et al.~\cite{DBLP:conf/nips/CranmerSBXCSH20} and Sanchez-Gonzalez et al.~\cite{Sanchez-Gonzalez2020}, hereon referred to as \textit{baseline 1}(BS 1) and \textit{baseline 2}(BS 2) respectively differ in the design of their encoder and processor. While their approaches are designed to predict the forward dynamics, we note that they are a special case of our approach such that removing the classification step, contact decoder and the inverse decoder from our approach tends to theirs. We consider them as baseline models to predict the forward dynamics. We find that BS 2 predicts the forward dynamics of elastic rigid-boddy collisions with an order of magnitude higher accuracy than BS 1. BS 2 features a parallel~(assuming unshared parameters) particle and edge model that computes hidden layer representation of particle and edge features that are concatenated to predict the next state of the system of particles. BS 1 on the other hand features a serial graph network wherein the edge features are transformed to hidden representations and then concatenated with raw particle features to predict the next state. As a consequence of this distinction, we find that when there are multiple particle type interactions, BS 2 predicts the next state with approximately an order of magnitude higher accuracy than BS 1. This can be observed by comparing their forward prediction errors on the 2D elastic collisions dataset. However, in the absence of multiple particle types, we find that BS 1 performs an order of magnitude better than BS 2. This difference shows that in the absence of particle types, encoded edge features play a dominant role in forward dynamics prediction. 

In an effort to unify current work such that they are agnostic to the presence of particle types and various particle interactions, we explore the following implicit biasing strategies. The processor in graph network models usually contain rich particle and edge information that we leverage by learning to predict 1) the contact matrix of particles and 2) particle type. In physics simulations involving direct elastic contacts~(collisions) between particles, every particle continues moving with their current velocity and acceleration until they experience a collision. We detect collisions between particles by computing a sigmoid activation on the elements of the contact matrix~($\hat{\textbf{C}}^t$). A value of 1 indicates collision between two particles while a value of 0 indicates absence of collision. We compute the activated sigmoid matrix and concatenate it with node-level features to then predict the next state. This computation introduces an inductive bias in the case of direct contact between elastic rigid-body particles that satisfies an implicit assumption that rigid-body simulators follow. From Table.~1~(2D elastic collisions), it is evident that implicit biasing improves the performance of predicting the next state of the particles across varying frame rate. Additionally, we compare the receiver operating characteristics score of the learned simulator across the various frame rates and find that the node classification accuracy is close to 96\% and remains consistent with change in frame-rates. In the absence of multiple particle-types and a system of directly colliding particles, we directly append raw particle features with relative particle features~($\hat{\textbf{C}}^t$). The 2D gravity with varying edge potential along with 2D spring like edge interaction experiments are several examples of systems that exhibit such phenomenon.

\subsubsection{Behavior of the inverse model}
\begin{figure}[!ht]
    \centering
    \includegraphics[width=0.5\textwidth]{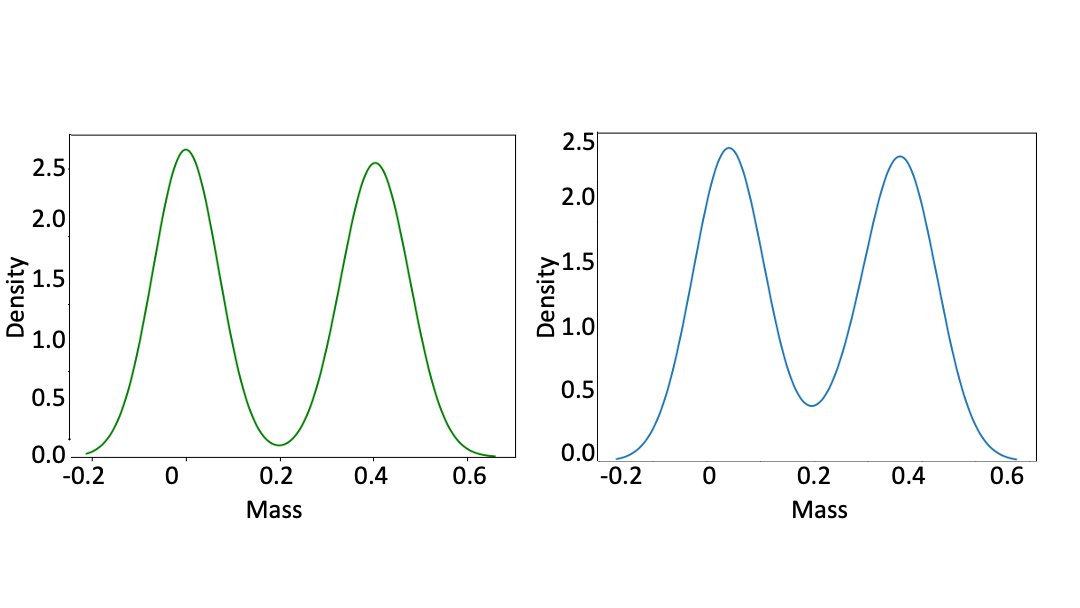}
    \caption{Mass distribution of particles subject to elastic rigid-body collisions.~[left: KDE of ground-truth mass distribution, right: KDE of approximate mass distribution]}
    \label{fig:Mass distribution of particles subject to elastic rigid-body collisions}
\end{figure}

\begin{figure}[!ht]
    \centering
    \includegraphics[width=0.5\textwidth]{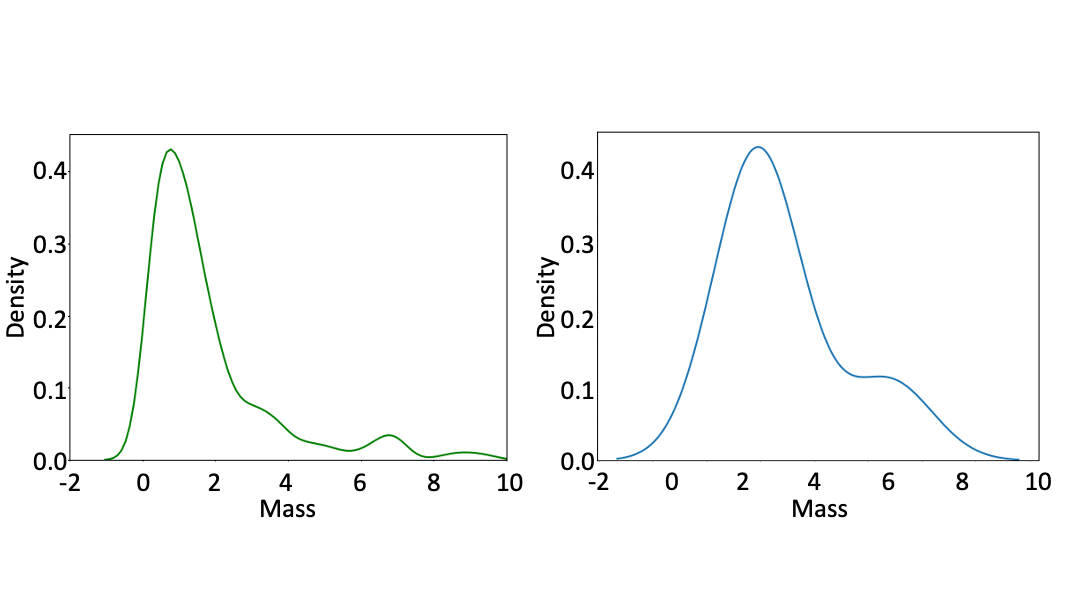}
    \caption{Mass distribution of particles subject to spring-force.~[left: KDE of ground-truth mass distribution, right: KDE of approximate mass distribution]}
    \label{fig:Mass distribution of particles subject to spring-force}
\end{figure}

Across our datasets, the unknown physical parameter of interest is the problem specific absolute particle mass. Each of the datasets pertaining to an analytic simulation model differs in particle mass distribution with varying complexity as shown by the ground-truth mass distribution. Figs.~\ref{fig:Mass distribution of particles subject to elastic rigid-body collisions} and \ref{fig:Mass distribution of particles subject to spring-force} show the ground-truth kernel density estimation of the mass distribution of particles along with their approximations. We observe that our inverse model has captured the modes~(0.001 and 0.4 in Fig. \ref{fig:Mass distribution of particles subject to elastic rigid-body collisions}) as well as the standard deviation of particle mass distributions with a high probability. We perform a t-test to compare the average values of the ground-truth mass distribution with the estimated mass distribution and we find with a p-value $> 0.8$ that the estimation of the inverse model and the ground-truth are not statistically significant across the datasets. While elastic rigid-body collision between particles comprises of a balanced dataset such that two modes that have similar density, particle masses subject to spring and gravitational forces are sampled from an imbalanced mass distribution with several low probabiility modes. While the approximation as shown by Fig.~\ref{fig:Mass distribution of particles subject to spring-force} covers the target mode, the approximation is much smoother and certain low probability regions have been assigned more support while some high probability regions have been assigned less support. We perform additional analysis~(please refer to appendix for more details) wherein we compute the $R^2$ value of the relative mass prediction using samples sampled from the posterior distribution of the relative mass distribution.

Finally, we note that we have used atleast 3 orders of magnitude fewer samples in comparison to the other works that we have bench-marked in this paper~\cite{DBLP:conf/nips/CranmerSBXCSH20,Sanchez-Gonzalez2020}. Similarly, compared with~\cite{DBLP:conf/uai/ZhengL0T18}, our graph-based approach requires atleast 1 order of magnitude fewer samples.
\section{Conclusion}
Currently, learnable physics simulators make use of explicit physics knowledge in the form of problem specific loss functions or implicit knowledge in the form of physics intuition and graph message-passing networks. In an effort to improve the functionality of existing graph networks we propose a data-driven graph network to simultaneously predict the forward state and the unknown physical parameter of a system of particles using multi-task learning objectives. We evaluate our approach on a diverse set of Newtonian physics models that simulate various interactions between particles. We explore biasing strategies such as particle classification and contact estimation, two related tasks that induce implicit learning of forward dynamics and inverse model. We find that our biasing strategies greatly improve the learning of forward dynamics while successfully inferring the probability distribution of an unknown particle specific physical parameter with orders of magnitude fewer samples. As future work, the proposed inverse model can be extended to discover multivariate as well as multiple particle-specific physical parameters. Finally, we highlight that our approach has considered an unweighted multi-task objective that equally gives importance to all tasks. This may not be ideal since all tasks may not carry equal importance and we speculate that additional objective function regularization might help the model converge to a better solution.

\section*{Acknowledgments}

\bibliography{AAAI_Submission_2022_Final_Copy}

\appendix
\section{Appendix}

\bigskip
\noindent 

\end{document}